\PassOptionsToPackage{noend}{algorithm2e}
\documentclass{l4dc2025}

\title[Contingency-MPPI]{Contingency Constrained Planning with MPPI within MPPI}
\usepackage{times}
\usepackage{floatrow}
\usepackage{float}
\usepackage{wrapfig}
\usepackage{caption}
\usepackage{color}
\usepackage{multicol}

\usepackage{caption}
\usepackage{subcaption}
\usepackage{tikz}
\usepackage[capitalize]{cleveref}

\newcommand\blfootnote[1]{%
  \begingroup
  \renewcommand\thefootnote{}\footnote{#1}%
  \addtocounter{footnote}{-1}%
  \endgroup
}

\newcommand{\lIfElse}[3]{\lIf{#1}{#2 \textbf{else}~#3}}

\usetikzlibrary{automata, positioning, arrows, decorations.pathreplacing, calc, chains, shapes.misc}

\coltauthor{%
 \Name{Leonard Jung} \Email{jung.le@northeastern.edu}\\
 \Name{Alexander Estornell} \Email{estornell.a@northeastern.edu}\\
 \Name{Michael Everett} \Email{m.everett@northeastern.edu}\\
 \addr Northeastern University, Boston, MA 02215, USA
}
\let\bm\boldsymbol
\LinesNumbered
\begin{document}

\maketitle
\blfootnote{Distribution Statement A. Approved for public release: distribution unlimited}
\blfootnote{\url{https://github.com/neu-autonomy/Contingency-MPPI}}
\begin{abstract}%
For safety, autonomous systems must be able to consider sudden changes and enact contingency plans appropriately. State-of-the-art methods currently find trajectories that balance between nominal and contingency behavior, or plan for a singular contingency plan; however, this does not guarantee that the resulting plan is safe for all time. To address this research gap, this paper presents Contingency-MPPI, a data-driven optimization-based strategy that embeds contingency planning inside a nominal planner.
By learning to approximate the optimal contingency-constrained control sequence with adaptive importance sampling, the proposed method's sampling efficiency is further improved with initializations from a lightweight path planner and trajectory optimizer.
Finally, we present simulated and hardware experiments demonstrating our algorithm generating nominal and contingency plans in real time on a mobile robot.\\
\textbf{Experiment Video} Video will be added soon
\end{abstract}


\section{Introduction}
Autonomous systems in real environments need to be able to handle sudden major changes in the operating conditions. For example, a car driving on the highway may need to swerve to safety if a collision occurs ahead, or a humanoid robot may need to grab hold of a railing if its foot slips on the stairs. Since there may be only a fraction of a second to recognize and respond to such events, this paper aims to develop an approach for always ensuring a contingency plan is available and can be immediately executed, if necessary.

A key challenge in this problem is to ensure a contingency plan always exists, without impacting the nominal plan too much. In standard approaches, where the nominal planner does not account for contingencies, the system could enter states from which no contingency plan exists; this may be tolerable if the failure event never occurs while the system is in one of those states but could lead to major safety failures in the worst case. \cite{9683388} considered these backup plans by adding weighted terms to the cost function, which \textit{encourage} staying out of these contingency-free areas, but does not have any guarantee. Alternatively, the methods of \cite{alsterda2021contingency} and \cite{9729171} explicitly plan alternative trajectories through a branching scheme and optimize both a nominal and set of contingency trajectories simultaneously. However, again, these methods must balance between the nominal trajectory cost and the tree of backup plans. \cite{9729171,li2023marc,10400882} all addressed risk-aware contingency planning with stochastic interactions with other agents; thus, these algorithms aim to minimize risk and cost over a tree of possible future scenarios, again only providing a balance between aggressive and safe behavior. \cite{tordesillas2019faster} solved a contingency constrained problem by planning both an optimistic plan and a contingency plan that branches off to stay in known-free space. However, for more complicated safety requirements (e.g., a collection of safe regions that must remain reachable within a given time limit), the mixed integer programming problem proposed in that work becomes too expensive to solve in real time.

This leads to another important challenge: ensuring computational efficiency despite needing to generate both a nominal and a contingency plan from each state along the way. One approach would be to use exact reachability algorithms \citep[e.g.,][]{8263977,9341499} to keep the planner out of contingency-free areas. However, computing the reachable sets is expensive, not strictly necessary, and would still require subsequently finding the contingency trajectories. Instead, this paper proposes to use an inexpensive contingency planner embedded inside the nominal planner. If the contingency planner finds a valid contingency plan, the nominal planner knows that state is acceptable, and the corresponding contingency plan is already available. Meanwhile, if the contingency planner fails to find a trajectory within a computation budget, the nominal planner can quickly (albeit conservatively) update its plans to avoid that state. To handle these discrete contingency events on top of generic nominal planning problems, our method, Contingency-MPPI, builds on model-predictive path integral (MPPI) control.

To summarize, this paper's contributions include:
(i) a planning algorithm that embeds contingency planning inside a nominal planner to ensure that a contingency plan exists from every state along the nominal plan,
(ii) extensions of this planner using lightweight optimization problems to improve the sample-efficiency via better initial guesses, and
(iii) demonstrations of the proposed method in simulated environments and on a mobile robot hardware platform to highlight the real-time implementation.

\section{Problem Formulation}
\begin{wrapfigure}{r}{0.42\textwidth}
\vspace{-0.8in}
    \centering    
    \includegraphics[width=\textwidth]{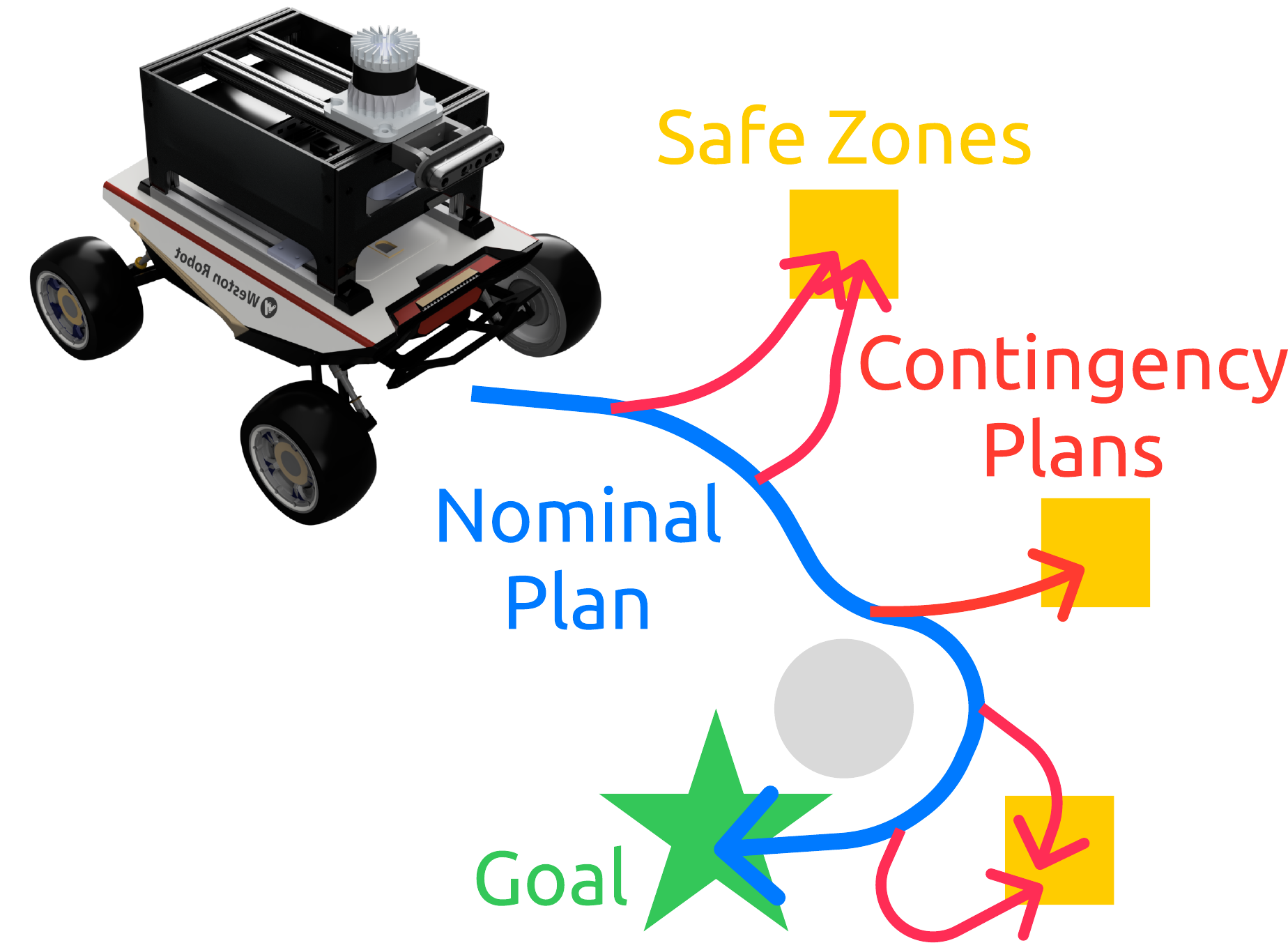}
    \vspace{-0.2in}
    \caption{At each step along the nominal plan, a contingency plan must exist to reach a safe state within a time horizon.}
    \label{fig:example_problem}
    \vspace{-0.2in}
\end{wrapfigure}

Denote a general nonlinear discrete-time system  $\mathbf{x}_{t+1}=f(\mathbf{x}_t, \mathbf{u}_t)$ with state, $\mathbf{x}_t\in\mathbb{R}^{n_x}$, and control, $\mathbf{u}_t \in\mathbb{R}^{n_u}$ at time $t$. To indicate a trajectory, we use colon notation (e.g., $\mathbf{x}_{0:T} = \{\mathbf{x}_0, \mathbf{x}_1, \ldots, \mathbf{x}_T\}$) and $T$-step dynamics $\mathbf{x}_{t+T} = f(\mathbf{x}_t, \mathbf{u}_{0:T})$. Additionally $\bm{U}\in \mathbb{R}^{n_uT}$ and $\bm{X} \in \mathbb{R}^{n_xT}$ indicate the state and control trajectory reshaped into a vector,  and $\mathbf{\Sigma}\in\mathbb{R}^{n_u T \times n_u T}$ represents the covariance matrix of the reshaped control trajectory.

The contingency-constrained planning problem is to find a nominal trajectory, $\mathbf{u}_{0:T}$, to minimize cost $J_{\text{nom}}$, along with contingency trajectories $\{\mathbf{u}^0_{0:T_c},\ \allowbreak\ldots,\ \allowbreak\mathbf{u}^T_{0:T_c}\}$ that drive the system into a safe set, $\mathcal{S}$, within $T_c$ steps, from each state along the nominal trajectory:
\begin{subequations}
\begin{align}
    \min_{\mathbf{u}_{0:T},\{\mathbf{u}^0_{0:T_c}, \ldots, \mathbf{u}^T_{0:T_c}\}} & J_\text{nom}(\mathbf{x}_0, \mathbf{u}_{0:T}) \\
    \text{s.t.}\quad\quad\quad& \mathbf{x}_{t+1} = f(\mathbf{x}_t, \mathbf{u}_t) &\forall{t \in [0,\ldots, T]}\,\,\\
    & \exists \tau\leq T_c\ \textrm{s.t.}\ f(\mathbf{x}_{i}, \mathbf{u}^i_{0:\tau}) \in \mathcal{S} &\forall{i \in [0,\ldots, T]}. \label{eq:reach_con}
\end{align}\label{eq:problem}
\end{subequations}
Other common costs/constraints (e.g., obstacle avoidance, control limits), can be added as desired.

\section{Contingency-MPPI}
\begin{figure}[htbp]
\centering
\includegraphics[width=12cm]{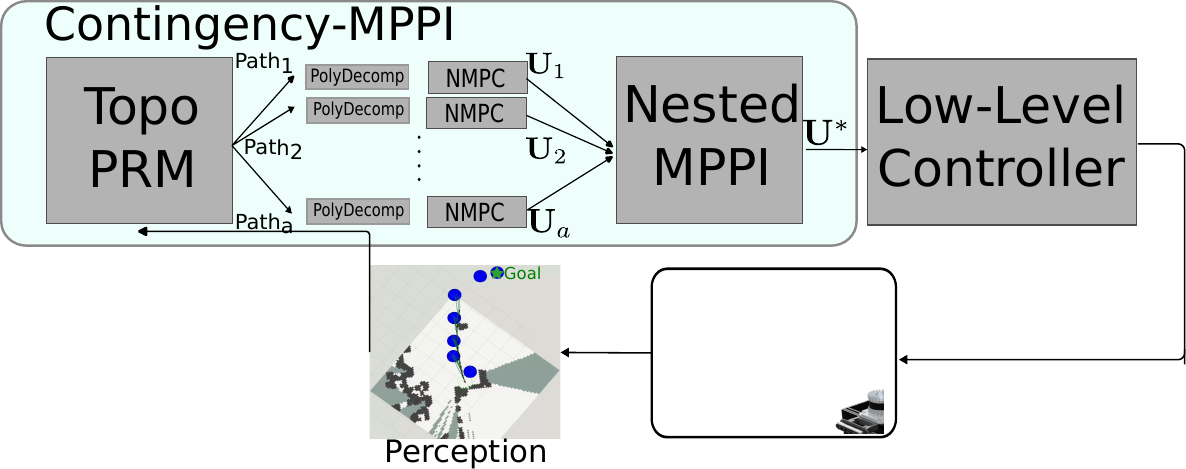}
\caption{Planning Pipeline. Our Contingency-MPPI first runs (1) TopoRPM to find multiple paths through the environment (\cref{sec:topoprm}), (2) NMPC to find control sequences for each path (\cref{sec:nmpc}), and (3) Nested-MPPI that utilizes these control sequences as modes (\cref{reachability_des}) to find a trajectory for the vehicle to track.}
\label{fig:architecture}
\end{figure}

Ignoring the contingency constraint \cref{eq:reach_con}, MPPI is a powerful method for solving \cref{eq:problem} when the dynamics or costs are non-convex.
This section shows how to extend MPPI to handle \cref{eq:reach_con} as well, by nesting another sampling process into MPPI.
First, we review the vanilla MPPI algorithm in \cref{MPPI_desc}, then describe our Nested-MPPI in \cref{reachability_des}. To increase sampling efficiency, a path-finding and trajectory optimization step (\cref{sec:frontend}) is used to seed Nested-MPPI (\cref{sec:backend}) with ancillary controllers. This approach is summarized in \cref{fig:architecture}.
\subsection{Background: Model Predictive Path Integral Control} \label{MPPI_desc}
To summarize MPPI following \cite{asmar2023model}, consider the entire control trajectory as a single input $\bm{V} \sim \mathcal{N}(\boldsymbol{U}, \boldsymbol{\Sigma})$, sampled from distribution $\mathbb{Q}$ with density
\begin{equation}
q(\boldsymbol{V})=((2\pi)^{n_uT}|\boldsymbol{\Sigma}|)^{-\frac{1}{2}}e^{-\frac{1}{2}(\boldsymbol{V}-\boldsymbol{U})^T\boldsymbol{\Sigma}^-1(\boldsymbol{V}-\boldsymbol{U})},
\end{equation} 
where $\bm{U},\bm{\Sigma}$ are the mean and covariance of $\mathbb{Q}$. The objective of MPPI is to minimize the KL-divergence between this proposed distribution, $\mathbb{Q}$, and an (unknown) optimal control distribution $\mathbb{Q}^*$, defined with respect to a cost function of the form
\begin{equation}
\mathcal{J}(\boldsymbol{X},\boldsymbol{U})=\mathbb{E_Q}[\phi(\boldsymbol{X})+c(\boldsymbol{X})+\frac{\lambda}{2}\boldsymbol{U}^T\boldsymbol{\Sigma}^{-1}\boldsymbol{U}].
\end{equation}
The optimal control distribution $\mathbb{Q}^*$ has density $q^*(\boldsymbol{V})=\frac{1}{\eta}(-\frac{1}{\lambda}S(\boldsymbol{V}))p(\boldsymbol{V})$, based on a state-dependent cost, $S(\boldsymbol{V})=\phi(\boldsymbol{X})+c(\boldsymbol{X})$ and a nominal control distribution, $\mathbb{P}$, with density
\begin{equation}
p(\boldsymbol{V})=((2\pi)^{n_uT}|\boldsymbol{\Sigma}|)^{-\frac{1}{2}}e^{-\frac{1}{2}(\boldsymbol{V}-\tilde{\boldsymbol{U}})^T\boldsymbol{\Sigma}^-1(\boldsymbol{V}-\tilde{\boldsymbol{U}})}.
\end{equation}
Here, $\eta$ is a normalizing constant, $\lambda$ is the inverse temperature, and $\tilde{\boldsymbol{U}}$ is the base control, which is usually either zero or a nominal distribution from iterations of adaptive importance sampling. Then, to find the optimal control trajectory we can minimize the KL-divergence between $\mathbb{Q}$ and $\mathbb{Q}^*$,
\begin{equation}
    \boldsymbol{U}^* =\arg\min_{\boldsymbol{U}}\mathbb{D}_{KL}(\mathbb{Q}^* || \mathbb{Q}).
\end{equation}
Using adaptive importance sampling, the optimal control can be approximated by drawing $N$ samples from a distribution $\mathbb{Q}_{\hat{\boldsymbol{U}}}$ with proposed input $\hat{\boldsymbol{U}}$,

\begin{eqnarray}
   & \boldsymbol{U}^* =\mathbb{E}_{\mathbb{Q}}[w(\boldsymbol{V})\boldsymbol{V}], 
    \label{eq:opt_u_input} \\
    & w(\boldsymbol{V})=\frac{1}{\eta}e^{-\frac{1}{\lambda}(S(\boldsymbol{V})+\lambda(\hat{\boldsymbol{U}}-\tilde{\boldsymbol{U}})^T\Sigma^{-1}\boldsymbol{V})}
    \\ & \eta=\int e^{-\frac{1}{\lambda}(S(\boldsymbol{V})+\lambda(\hat{\boldsymbol{U}}-\tilde{\boldsymbol{U}})^T\Sigma^{-1}\boldsymbol{V})} d\boldsymbol{V}.
    \vspace{-1cm}
\end{eqnarray}
(\ref{eq:opt_u_input}) finds an (information-theoretic) optimal open-loop control sequence that can be implemented in a receding horizon by shifting one timestep ahead and re-running the algorithm.
As in \cite{asmar2023model}, we weigh the control cost by a factor $\gamma=\lambda(1-\alpha)$ and shift all sampled trajectory costs by the minimum sampled cost, for numerical stability.
\subsubsection{Enforcing Hard Constraints in MPPI}\label{pseudho_hard_des}
While MPPI does not explicitly handle constraints, such as avoiding obstacles or the existence of a contingency plan, one can add terms to the objective with infinite cost when constraints are violated. For example, with a nominal cost $S_{\text{nom}}(\bm{V})$ and $N$ constraints, the augmented cost is $S(\boldsymbol{V})=S_{\text{nom}}(\bm{V})+\sum_{k=1}^{N} S_{\text{constraint}}^k(\bm{V})$, where 
\begin{equation}
    S_{\text{constraint}}^k(\bm{V})=\begin{cases}
        0, & \text{if constraint $k$ is satisfied}\\
        \infty, & \text{o.w.}
        \end{cases}.
\end{equation}

When the constraints are satisfied, the trajectory cost (and its weight in importance sampling) depends only on the nominal cost, such as minimum time or distance to the goal. If the constraints are not satisfied, the trajectory has infinite cost and receives zero weight in importance sampling. Thus, only trajectories meeting all constraints are considered, and their weights depend solely on the nominal cost. If no trajectory satisfies all constraints in an iteration, all samples get zero weight, and the mean control trajectory remains unchanged for the next MPPI iteration. However, if any trajectory met the constraints in the previous step, the executed control trajectory will still be safe

\subsection{Nested-MPPI}\label{reachability_des}
\begin{figure}[htpb]
\centering
\includegraphics[width=0.55\textwidth]{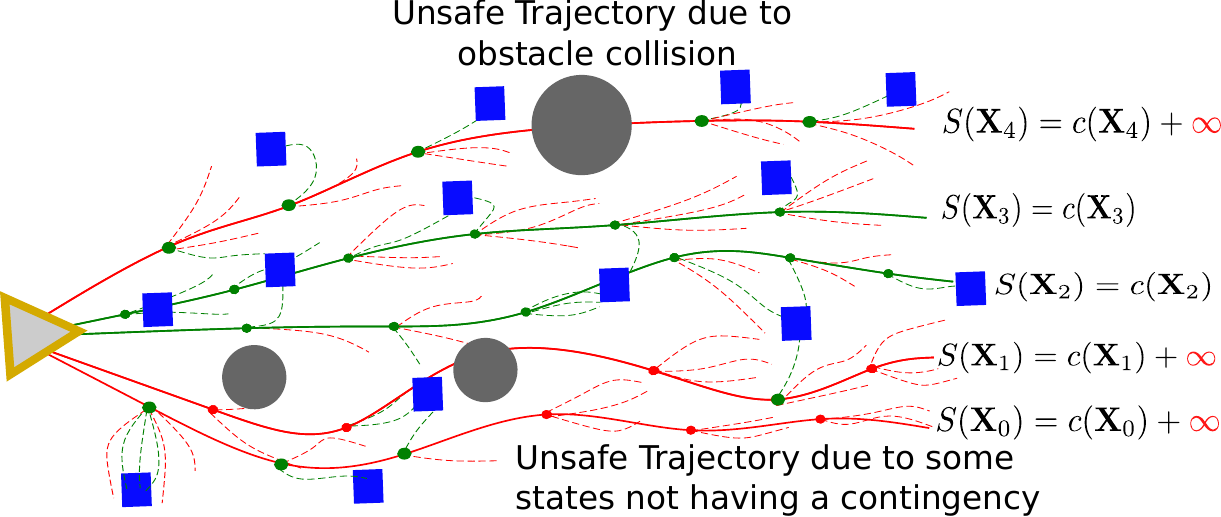}
\caption{Nested-MPPI computes reachability cost by sampling contingency trajectories (dashed lines) along nominal trajectory rollouts (solid lines). Nominal trajectories 0, 1, and 4 collided with an obstacle or did not find a valid contingency from every state, and thus have $+\infty$ cost.}
\label{fig:branching_explanation}
\end{figure}
\begin{minipage}[t]{0.48\textwidth}
    \begin{algorithm2e}[H]
\DontPrintSemicolon

\textbf{Input:} $\mathbf{x}_0, \bm{U}, [\bm{U}_a], \bm{\Sigma}$\\
\textbf{Output:} Nominal \& contingency control sequences\\
\textbf{Parameters:} $K, T, L$ (nominal); $f, G, $ (system); $c, \phi, \lambda, \alpha$ (cost)\\
\BlankLine

$\bm{U}' \gets \bm{U}$; $\bm{\Sigma}' \gets \bm{\Sigma}$\;
\For(\tcp*[f]{AIS loop}){$l \gets 0$ \KwTo $L-1$} { 
    \For{$k \gets 0$ \KwTo $K-1+\text{card}([\bm{U}_a])$}{ \label{start_jax_batch_3}
        $\mathbf{x}_{k,0} \gets \mathbf{x_0}$\\
        $\bm{\mathcal{E}}_k \sim \mathcal{N}(0, \bm{\Sigma'})$\\
        \lIf{$k \leq K-1$}{$\bm{U}=\bm{U}'+\bm{\mathcal{E}}_k$ } \label{line:start_anci} \tcp{Ancillary Control}
        \lElse{$\bm{U}=[\bm{U}_a]_{k-(K-1)}$} \label{line:end_anci} 
        \For{$i \gets 0$ \KwTo $T-1$}{
            $\mathbf{x}_{k,i+1} = \mathbf{x}_{k,i} + \left( f + G \left(\mathbf{u} \right) \right) \Delta t$
        }
        $\bm{U}_{\text{reach}_k}, S_{\text{reach}} \gets \text{FindContingencyPlan}(\bm{X}_k)$\;
        \label{line:findcontingencyplan}
        \lIf{$S_{\text{reach}} = 0$}{ $\bm{U}_{\text{contingency}} \gets \bm{U}_{\text{reach}_k}$}
        $S_k \gets S_{\text{reach}} + c(\bm{X}) + \phi(\bm{X}) + \lambda(1-\alpha)\bm{U}'^T\bm{\Sigma}^{-1}(\bm{\mathcal{E}}_k + \bm{U}' - \bm{U})$
    }\label{end_jax_batch_3}
    \lIf{$l < L - 1$}{ $\bm{U}', \bm{\Sigma}' \gets \text{AIS}()$}
}

\BlankLine
$\rho \gets \min(\textbf{S})$\;
$\eta \gets \sum_{k=1}^K \exp\left( -\frac{1}{\lambda}(S_k - \rho)\right)$

\For{$k \gets 0$ \KwTo $K-1$}{
    $\bm{U} \mathrel{+}= \frac{1}{\eta}\exp\left( -\frac{1}{\lambda}(S_k - \rho)\right) \left( \bm{\mathcal{E}}_k + \bm{U}' - \bm{U} \right)$
}

\Return $\bm{U}, \bm{U}_{\text{contingency}}$
\caption{Nested-MPPI}
\label{alg:nested}
\end{algorithm2e}
\end{minipage}%
\begin{minipage}[t]{0.52\textwidth}
    \begin{algorithm2e}[H]
\DontPrintSemicolon

\textbf{Input:} $\bm{X}$: State sequence\\
\textbf{Output:} Contingency control sequence \& score\\
\textbf{Parameters:} $K_{\text{c}}, T_{\text{c}}, L_{\text{c}}, T_{\text{s}}$ (contingency);
       $f, G, m_{\text{elite}}$ (system/sampling);
       $\varepsilon, \lambda, \alpha$ (costs)\\
\BlankLine
$\bm{U}' \leftarrow \bm{0}$

\For{$i \gets 0$ \KwTo $T_{\text{s}}-1$}{ \label{start_jax_batch_1}
    \For{$l \gets 0$ \KwTo $L_{\text{c}}-1$}{ \label{start_of_sampling}
        $\mathbf{x} \gets \mathbf{x}_i$\;
        \For{$k \gets 0$ \KwTo $K_{\text{c}}-1$}{\label{start_jax_batch_2}
            \lIfElse{$l = 0$}{$\bm{\mathcal{E}}_k \sim \text{U}(\bm{u}_{lb}, \bm{u}_{ub})$}{$\bm{\mathcal{E}}_k \sim\mathcal{N}(0, \bm{\Sigma})$} 
            \For{$i \gets 0$ \KwTo $T-1$}{
                $\mathbf{x}_{k,i+1} = \mathbf{x}_{k,i} + (f + G(\mathbf{u}_i' + \bm{\epsilon}_{i,k}')) \Delta t$
            }
            $S_k \gets \min_{	\zeta  \in \mathcal{S}, \mathbf{x} \in \boldsymbol{X}^s} \|\mathbf{x} - 	\zeta \|$\;
            \lIf{$l > 0$}{$S_k \mathrel{+}= \lambda(1-\alpha)\bm{U}'^T \bm{\Sigma}^{-1}(\bm{\mathcal{E}}_k + \bm{U}' - \bm{U})$}
        }\label{end_jax_batch_2}
        \lIf{$l = 0$}{ \label{start_initialize_dist}
            $\bm{U}_\text{best} \gets \text{selectBest}(\bm{U}, m_\text{elite}, S)$
            $\bm{U}', \bm{\Sigma}' \leftarrow \text{Mean}(\bm{U}_\text{best}), \text{Cov}(\bm{U}_\text{best})$
        }

        \lElseIf{$l < L_{\text{c}}-1$}{
            $\bm{U}', \bm{\Sigma}' \leftarrow \text{AIS}()$
        } 
    }\label{end_of_sampling}
    \lIfElse{$\min(c_i) < \varepsilon$}{$S_{i,\text{reach}} \gets 0$}{$S_{i,\text{reach}} \gets \infty$} 

} \label{end_jax_batch_1}

\Return $\bm{U}'_0, \sum_{i=0}^{T_{\text{safe}}-1} S_{i,\text{reach}}$
\caption{$\text{FindContingencyPlan}$}
\label{alg:reachr}
\end{algorithm2e}
\end{minipage}

This section introduces Nested-MPPI, which is summarized in \cref{alg:nested}. This algorithm is based on the MPPI in \cite{alsterda2021contingency} and allows for ancillary controllers as proposed in \cite{trevisan2024biased}. Our key innovation begins on \cref{line:findcontingencyplan}, where a second level of MPPI executes in each state along each rollout of the nominal MPPI.
This second level, described in \cref{alg:reachr}, optimizes for a contingency plan (with a different cost function than the nominal plan) as a way of evaluating the reachability constraint, \cref{eq:reach_con}.

To both find contingency trajectories and evaluate whether the reachability constraint (\ref{eq:reach_con}) is satisfied for any control sequence, we first roll out each control sequence $\bm{U}_i$ by passing it through the zero-noise nonlinear dynamics model to get the state sequence $\boldsymbol{X}_i$. Then, at each state $\mathbf{x}_t$ for $t=0,\dots,T_{s}-1$, we run $L_{\text{c}}$ rounds of adaptive importance sampling MPPI with $N_{c}$ samples and $T_{c}$ timesteps starting at $\mathbf{x}_t$ to generate contingency state $[\bm{X}_0^s,\dots,\bm{X}_{L\times N_c}^s]$ and control $[\bm{U}_0^s,\dots,\bm{U}_{L\times N_c}^s]$ trajectories (lines \ref{start_of_sampling}-\ref{end_of_sampling} in Alg. \ref{alg:reachr}). To encourage these contingency trajectories to reach safe states within $T_{c}$ timesteps, we use state-dependent cost
\begin{equation}
    c_\text{contingency}(\boldsymbol{X}^s) = \min_{	\zeta \in \mathcal{S}, \mathbf{x}\in\boldsymbol{X}^s}\|\mathbf{x}-	\zeta \|.
\end{equation}

Then, as seen in Figure \ref{fig:branching_explanation}, if any of the contingency trajectories successfully reaches a safe state within $T_{c}$ timesteps, we mark that state $\mathbf{x}_t$ along $\bm{X}_i$ as safe (green). If all states along $\bm{X}_i$  are marked as safe, then we mark $\bm{X}_i$ and its corresponding control $\bm{U}_i$ as safe; otherwise, we mark the trajectory as unsafe, and add $+\infty$ to its corresponding cost. In \cref{alg:reachr} \cref{start_initialize_dist}, to initialize the proposed distribution for contingencies at each state, a number of sample control trajectories are drawn from a uniform distribution along control bounds $\mathbf{u}_{t,0:T_{c}} \sim U(\mathbf{u}_{\text{lb}}, \mathbf{u}_\text{ub})$, and the cross-entropy method is used on the best $m$ trajectories to determine an initial mean and covariance.

\subsection{Improving the Sampling Efficiency of Nested-MPPI: Frontend} \label{sec:frontend}
Although \cref{alg:nested} considers all costs and constraints from \cref{eq:problem}, the sampling process can result in many or even all trajectories with infinite costs (if the sampling distribution is far from the optimal distribution), which leads to uninformed updates to the distribution. To remedy this, one may simply sample more $\boldsymbol{U}$ sequences; however, each additional sequence requires computing $S_{\text{reach}}$, which requires an additional $T_{s}$ MPPI computations. Thus, rather than simply increasing the number of samples, we propose to approximate locally optimal $\boldsymbol{U}$ and consider them as a new sampling distribution(s) into \cref{alg:nested}, as described in \cref{sec:backend}. First, we find several different paths between the start and goal. For each path, we then perform a convex decomposition to find an under approximation of the safe space, and finally perform a nonlinear MPC to solve for a candidate control sequence.

\subsubsection{Topo-PRM} \label{sec:topoprm}
To find several alternative paths through the workspace, we leverage Topo-PRM proposed in \cite{9196996}. As Topo-PRM finds a collection of topologically distinct paths through the environment, our planner can "explore" the free-space and return multiple promising guiding paths. However as Topo-PRM does not consider safe zones, it may return paths that are not near safe zones, and thus no contingencies exist. Thus, we modify the algorithm to sample randomly from safe states $p$ fraction of the time to bias our roadmap to find paths that include safe states.
To further bias the paths towards safe zones, denoting $V_\text{max}$ as the maximum speed of our vehicle in the creation of our workspace occupancy grid, we add pseudo-obstacles by marking occupied voxels that are farther than $r_\text{max}=V_\text{max}T_{s}\Delta t$ away from any given safe zone.

\subsubsection{Nonlinear MPC}\label{sec:nmpc}

To transform each path into an ancillary control trajectory, we first find the point $E$ that is $r_\text{max}$ along the path, and then perform a convex decomposition using the approach from \cite{7839930} of the free space along the path from our start point $S$ to $E$. Next, we find $M$ knot points by discretizing $M-1$ points along the path from $S$ to $E$. Denoting $A_i \mathbf{p}<b_i$ the polyhedral constraint in which point $p_i$ lies within, we solve the following nonlinear programming problem to recover a candidate ancillary control trajectory:

\begin{align} \label{nmpc}
\begin{split}
     U_{\text{anci}}=\arg\min_{\mathbf{X},\mathbf{U}}&\sum_{i=0}^{N} \| \mathbf{x}_{i}-\mathbf{x}_{goal} \|
    \\\text{s.t.}&\quad A_{i}\beta \mathbf{x}_{i}<b_{i}
    \\&\quad \mathbf{x}_{i+1}=f(\mathbf{x}_{i},\mathbf{u}_{i})
    \\&\quad \mathbf{x}_{lb}<\mathbf{x}_{i}<\mathbf{x}_{ub}
    \\&\quad \mathbf{u}_{lb}<\mathbf{u}_{i}<\mathbf{u}_{ub}
    \\&\quad \qquad \forall i\in\{0,M\}
    \\&\quad \mathbf{x}_{0}=\mathbf{x}_{start},
\end{split}
\end{align}
where $\beta \in \mathbb{R}^{n_x \times n_x}$ selects elements of the state space present in the workspace (i.e., $\mathbf{p}_i = \beta \mathbf{x}_i$).

\begin{wrapfigure}{r}{0.5\textwidth}
    \begin{minipage}{0.95\textwidth}
      \vspace{-0.2in}
      \input{alg_seeded_mppi}
      \vspace{-0.4in}
    \end{minipage}
\end{wrapfigure}
\subsection{Improving the Sampling-Efficiency of Nested-MPPI: Backend} \label{sec:backend}
To use each of the control sequences found in \cref{sec:frontend}, we sample around the control sequences and insert the resulting set of samples as biases to MPPI as in \cite{trevisan2024biased}.
As seen in Alg.  \ref{alg:con-mppi}, we append each of the $B$ control sequences from MPC to a list of ancillary control sequences. Then, in Alg. \ref{alg:nested}, the control sequence is used as a mean for an individual distribution, $\bm{\mathcal{V}} \sim \mathcal{N}(\bm{U}_{\text{anci}}, \bm{\Sigma})$, where $\boldsymbol{\Sigma}$ is the current covariance of our MPPI algorithm, and sample $N_a$ samples. For the rest of the $N-B \cdot N_{a}$ samples the current MPPI running mean is used (\cref{alg:nested} Lines \ref{line:start_anci}-\ref{line:end_anci}). Intuitively, this is equivalent to inserting each bias as a mode in a Gaussian mixture model.

\section{Experiments}
 We demonstrate Contingency-MPPI on a hide-and-seek task in both simulation and hardware to highlight it's ability to (1) ensure a contingency plan exists at all time and (2) run in real-time within an autonomy stack. Critically, our results show that Contingency-MPPI not only guarantees the existence of contingencies but computes the trajectories as well. At a moment's notice, our planner can switch to its contingency behavior without needing to replan a new trajectory.

\subsection{Implementation Details}
As discussed in \cite{asmar2023model}, multiple adaptive importance sampling methods may be used; we chose to use the cross-entropy method for its simplicity and speed. To enable real-time computation, each iteration of lines \ref{start_jax_batch_3}-\ref{end_jax_batch_3} in Alg. \ref{alg:nested} is batched using JAX \citep{jax2018github}. Additionally, each iteration of lines \ref{start_jax_batch_1}-\ref{end_jax_batch_1} and \ref{start_jax_batch_2}-\ref{end_jax_batch_2} in Alg. \ref{alg:reachr} are also batched. Thus, the Nested-MPPI algorithm run time scales roughly by $L(T+L_{s}(T_c))$. For longer time horizons, we treated the control sequence and state sequence as a reference trajectory for a lower-level controller \citep[iLQG,][]{1469949} to track.
To solve the shooting NMPC problem, we used CasADi \citep{Andersson2019} and SNOPT \citep{10.1137/S0036144504446096}.

\subsection{Hide and Seek}
The experimental environment consists of several safe positions $p=[x,y]$, a start pose, an end position, and several obstacles. The objective is to control a differential-drive car from the start pose to the end position in minimum time, while satisfying the reachability constraint. Thus, the nominal cost is
\begin{equation}
    c_{\text{nom}}(\bm{X})=\sum_{t=0}^{T-1}{(\mathbf{x}_t-\mathbf{x}_{\text{goal}})^T\boldsymbol{Q}(\mathbf{x}_t-\mathbf{x}_{\text{goal}})}.
\end{equation}

We assume the location of safe positions is known, but the robot has a limited sensing horizon and thus operates with unknown obstacles. To handle this, an additional constraint is added to the cost function:
\begin{equation}
    c_{\text{known}}(\bm{X})=\begin{cases}
        0 & \mathbf{x} \ \text{exists in known space} \ \forall t\in\{0,\dots,T_{\text{safe}}-1\}\\
        \infty & \text{o.w}
        \end{cases},
\end{equation}
which ensures all states $i=0,\dots,T_{\text{safe}}$ in the nominal trajectories remain in the known region. 

\subsection{Simulation Results}

\begin{figure}[ht]
\RawFloats
    \centering
    \includegraphics[width=0.7\linewidth]{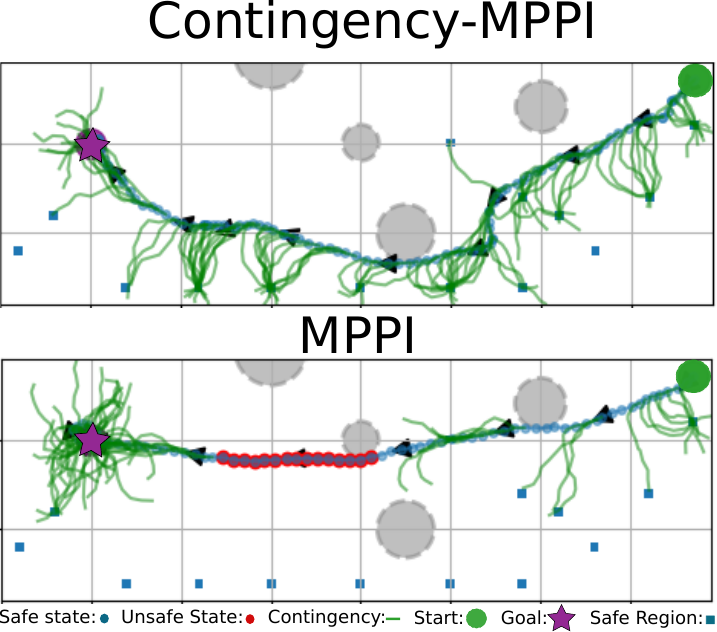}
    \caption{Example run from the simulation environment, showing the path the robot traversed and a single \textcolor{green}{contingency trajectory} at each state that would have been able to reach a \textcolor{blue}{safe state} within $T_c$ timesteps. While Contingency-MPPI guides the nominal plans such that a contingency plan always exists, MPPI drives the systems into \textcolor{red}{unsafe states}.}
    \label{fig:sim}
    \includegraphics[width=0.7\linewidth]{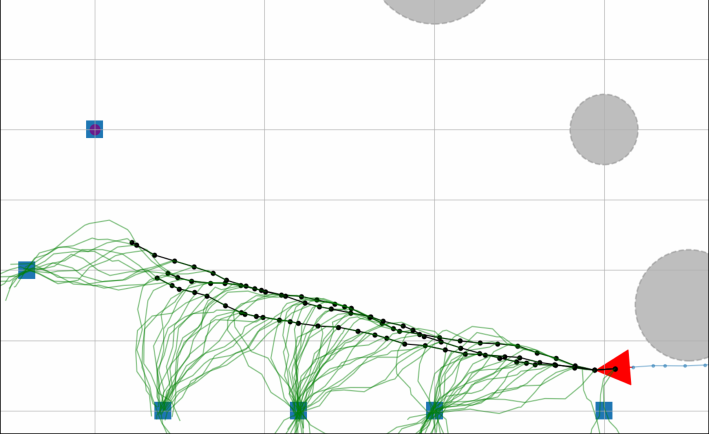}
    \captionof{figure}{Zooming into a particular timestep of Contingency-MPPI, there may be several nominal rollouts for which valid contingencies exist for \emph{every timestep} (3 such rollouts are shown here).}
    \label{fig:sim2}
\end{figure}

\begin{table}[tbp]
\centering
\begin{tabular}{|c|ccc|c|}
\hline
& \multicolumn{3}{c|}{Contingency-MPPI Variants} & \\
& AIS-MPC & MPC  & Base & MPPI \\ \hline
Reached Goal Rate (\%) $\uparrow$    & 92.0    & 69.3 & 44.0 & 100.0\\ \hline
Unsafe States (\%) $\downarrow$        & 0.0     & 0.0  & 0.0  & 4.7  \\ \hline
Average Timestep To Goal $\downarrow$  & 62.9    & 63.1 & 78.6 & 61.5 \\ \hline
Finite Cost Sampling (\%) $\uparrow$ & 53.7    & 55.0 & 47.4 & N/A \\ \hline
\end{tabular}
\caption{Simulation Results over 150 environments. As each environment was randomly generated, a trajectory to the goal with contingency trajectories is not guaranteed to exist.}
\label{table:1}
\end{table}

We ran 3 variants of our algorithm against a set of 150 randomly generated problems: the MPPI algorithm is the original MPPI algorithm considering only the nominal cost, and not the reachability safety constraint; Base algorithm is the Nested-MPPI algorithm; MPC includes MPC seeding with Nested-MPPI; AIS-MPC additionally considers adaptive importance sampling for the branching contingency control trajectories. The proposed methods always provide a contingency path to a safe zone anywhere along its trajectory (0\% unsafe states), while MPPI alone enters unsafe states. Additionally, \cref{table:1} highlights the advantage of using the topological NMPC frontend and AIS branched Contingency-MPPI in terms of successful solve, given the same sampling parameters. An example run in \cref{fig:sim}  compares AIS-MPC and vanilla MPPI. Our Contingency-MPPI always has a contingency trajectory to a safe zone, while the vanilla MPPI violates this safety condition for several time steps.
\cref{fig:sim2} provides a closer look at the rollouts under consideration at a particular timestep of AIS-MPC.

\subsection{Hardware Results}
\begin{figure}[tbp]
 \includegraphics[width=\textwidth]{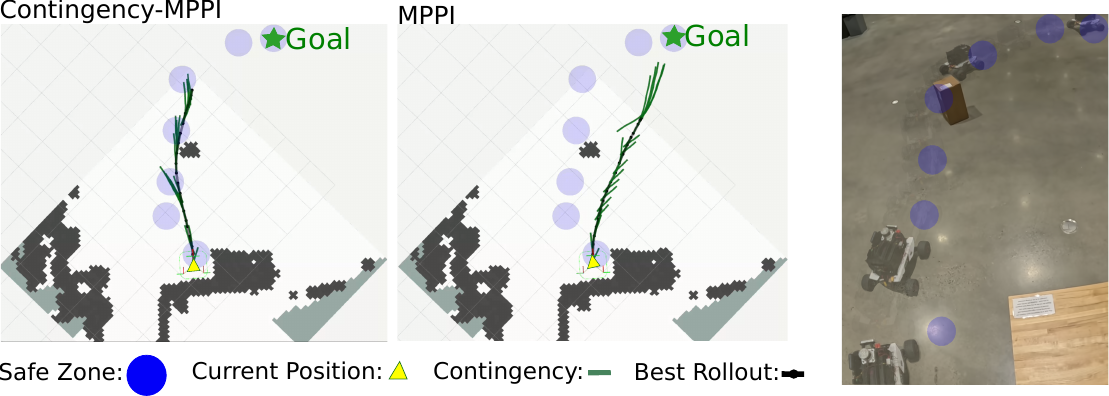}
    {
    \caption{Example from Hardware Experiment. (Left) Best rollout for Contingency-MPPI vs MPPI at $t=0$ with contingency trajectories at each timestep. (Right) Snapshot view of robot executing Contingency-MPPI.}
    \label{fig:hardware}
    }
\end{figure}

For the hardware experiments, we ran Contingency-MPPI on an Alienware Intel NUC 11 with an i7 processor and RTX 2060 GPU onboard our customized Agile-X Scout Mini platform with an Ouster OS1-32 3D lidar sensor. For robot localization and occupancy map creation, we ran direct lidar-inertial odometry \citep[DLIO,][]{10160508} on the same computer as the planner. A frame from an example trial is shown in \cref{fig:hardware} with video results in the supplementary material. Contingency-MPPI provides contingencies to the safe zones at every timestep along the nominal plan, while vanilla MPPI's rollout does not. The hardware experiments were conducted at varying speeds from 1-3 m/s with a varying number of obstacles, and all trials were conducted without a prior obstacle map; thus, our algorithm was capable of running in unknown environments and in real time.

\section{Summary}
This paper presented a method to solve contingency-constrained planning problems, enforcing the existence of a finite-time contingency trajectory at all timesteps along our robot's planned trajectory. The proposed algorithm, Contingency-MPPI, handles this constraint with adaptive importance sampling along each control trajectory. To improve sampling efficiency, the approach includes a separate MPC process to seed the MPPI modes. Finally, we demonstrate Contingency-MPPI both in simulation and hardware. Future work will consider high-dimensional dynamics, where we expect the required sampling density to greatly increase. Similarly, we plan to investigate incorporating other types of contingency behaviors, such as lane weaving on a highway or staying close to a shoulder along a road.
\acks{This work was supported by award W911NF-24-2-006.}

\bibliography{refs}

\begin{thebibliography}{17}
\providecommand{\natexlab}[1]{#1}
\providecommand{\url}[1]{\texttt{#1}}
\expandafter\ifx\csname urlstyle\endcsname\relax
  \providecommand{\doi}[1]{doi: #1}\else
  \providecommand{\doi}{doi: \begingroup \urlstyle{rm}\Url}\fi

\bibitem[Alsterda and Gerdes(2021)]{alsterda2021contingency}
John~P Alsterda and J~Christian Gerdes.
\newblock Contingency model predictive control for linear time-varying systems.
\newblock \emph{arXiv preprint arXiv:2102.12045}, 2021.

\bibitem[Andersson et~al.(2019)Andersson, Gillis, Horn, Rawlings, and Diehl]{Andersson2019}
Joel A~E Andersson, Joris Gillis, Greg Horn, James~B Rawlings, and Moritz Diehl.
\newblock {CasADi} -- {A} software framework for nonlinear optimization and optimal control.
\newblock \emph{Mathematical Programming Computation}, 11\penalty0 (1):\penalty0 1--36, 2019.
\newblock \doi{10.1007/s12532-018-0139-4}.

\bibitem[Asmar et~al.(2023)Asmar, Senanayake, Manuel, and Kochenderfer]{asmar2023model}
Dylan~M Asmar, Ransalu Senanayake, Shawn Manuel, and Mykel~J Kochenderfer.
\newblock Model predictive optimized path integral strategies.
\newblock In \emph{2023 IEEE International Conference on Robotics and Automation (ICRA)}, pages 3182--3188. IEEE, 2023.

\bibitem[Bansal et~al.(2017)Bansal, Chen, Herbert, and Tomlin]{8263977}
Somil Bansal, Mo~Chen, Sylvia Herbert, and Claire~J. Tomlin.
\newblock Hamilton-jacobi reachability: A brief overview and recent advances.
\newblock In \emph{2017 IEEE 56th Annual Conference on Decision and Control (CDC)}, pages 2242--2253, 2017.
\newblock \doi{10.1109/CDC.2017.8263977}.

\bibitem[Bradbury et~al.(2018)Bradbury, Frostig, Hawkins, Johnson, Leary, Maclaurin, Necula, Paszke, Vander{P}las, Wanderman-{M}ilne, and Zhang]{jax2018github}
James Bradbury, Roy Frostig, Peter Hawkins, Matthew~James Johnson, Chris Leary, Dougal Maclaurin, George Necula, Adam Paszke, Jake Vander{P}las, Skye Wanderman-{M}ilne, and Qiao Zhang.
\newblock {JAX}: composable transformations of {P}ython+{N}um{P}y programs, 2018.
\newblock URL \url{http://github.com/jax-ml/jax}.

\bibitem[Chen et~al.(2023)Chen, Nemiroff, and Lopez]{10160508}
Kenny Chen, Ryan Nemiroff, and Brett~T. Lopez.
\newblock Direct lidar-inertial odometry: Lightweight lio with continuous-time motion correction.
\newblock In \emph{2023 IEEE International Conference on Robotics and Automation (ICRA)}, pages 3983--3989, 2023.
\newblock \doi{10.1109/ICRA48891.2023.10160508}.

\bibitem[Chen et~al.(2022)Chen, Rosolia, Ubellacker, Csomay-Shanklin, and Ames]{9729171}
Yuxiao Chen, Ugo Rosolia, Wyatt Ubellacker, Noel Csomay-Shanklin, and Aaron~D. Ames.
\newblock Interactive multi-modal motion planning with branch model predictive control.
\newblock \emph{IEEE Robotics and Automation Letters}, 7\penalty0 (2):\penalty0 5365--5372, 2022.
\newblock \doi{10.1109/LRA.2022.3156648}.

\bibitem[Gill et~al.(2005)Gill, Murray, and Saunders]{10.1137/S0036144504446096}
Philip~E. Gill, Walter Murray, and Michael~A. Saunders.
\newblock Snopt: An sqp algorithm for large-scale constrained optimization.
\newblock \emph{SIAM Rev.}, 47\penalty0 (1):\penalty0 99–131, jan 2005.
\newblock ISSN 0036-1445.
\newblock \doi{10.1137/S0036144504446096}.
\newblock URL \url{https://doi.org/10.1137/S0036144504446096}.

\bibitem[Kim et~al.(2021)Kim, Yoon, Wan, Hovakimyan, Sha, and Voulgaris]{9683388}
Hunmin Kim, Hyungjin Yoon, Wenbin Wan, Naira Hovakimyan, Lui Sha, and Petros Voulgaris.
\newblock Backup plan constrained model predictive control.
\newblock In \emph{2021 60th IEEE Conference on Decision and Control (CDC)}, pages 289--294, 2021.
\newblock \doi{10.1109/CDC45484.2021.9683388}.

\bibitem[Li et~al.(2023)Li, Zhang, Liu, and Shen]{li2023marc}
Tong Li, Lu~Zhang, Sikang Liu, and Shaojie Shen.
\newblock Marc: Multipolicy and risk-aware contingency planning for autonomous driving.
\newblock \emph{IEEE Robotics and Automation Letters}, 2023.

\bibitem[Liu et~al.(2017)Liu, Watterson, Mohta, Sun, Bhattacharya, Taylor, and Kumar]{7839930}
Sikang Liu, Michael Watterson, Kartik Mohta, Ke~Sun, Subhrajit Bhattacharya, Camillo~J. Taylor, and Vijay Kumar.
\newblock Planning dynamically feasible trajectories for quadrotors using safe flight corridors in 3-d complex environments.
\newblock \emph{IEEE Robotics and Automation Letters}, 2\penalty0 (3):\penalty0 1688--1695, 2017.
\newblock \doi{10.1109/LRA.2017.2663526}.

\bibitem[Peters et~al.(2024)Peters, Bajcsy, Chiu, Fridovich-Keil, Laine, Ferranti, and Alonso-Mora]{10400882}
Lasse Peters, Andrea Bajcsy, Chih-Yuan Chiu, David Fridovich-Keil, Forrest Laine, Laura Ferranti, and Javier Alonso-Mora.
\newblock Contingency games for multi-agent interaction.
\newblock \emph{IEEE Robotics and Automation Letters}, 9\penalty0 (3):\penalty0 2208--2215, 2024.
\newblock \doi{10.1109/LRA.2024.3354548}.

\bibitem[Todorov and Li(2005)]{1469949}
E.~Todorov and Weiwei Li.
\newblock A generalized iterative lqg method for locally-optimal feedback control of constrained nonlinear stochastic systems.
\newblock In \emph{Proceedings of the 2005, American Control Conference, 2005.}, pages 300--306 vol. 1, 2005.
\newblock \doi{10.1109/ACC.2005.1469949}.

\bibitem[Tordesillas et~al.(2019)Tordesillas, Lopez, and How]{tordesillas2019faster}
Jesus Tordesillas, Brett~T Lopez, and Jonathan~P How.
\newblock Faster: Fast and safe trajectory planner for flights in unknown environments.
\newblock In \emph{2019 IEEE/RSJ international conference on intelligent robots and systems (IROS)}, pages 1934--1940. IEEE, 2019.

\bibitem[Trevisan and Alonso-Mora(2024)]{trevisan2024biased}
Elia Trevisan and Javier Alonso-Mora.
\newblock Biased-mppi: Informing sampling-based model predictive control by fusing ancillary controllers.
\newblock \emph{IEEE Robotics and Automation Letters}, 2024.

\bibitem[Wang et~al.(2020)Wang, Leung, and Pavone]{9341499}
Xinrui Wang, Karen Leung, and Marco Pavone.
\newblock Infusing reachability-based safety into planning and control for multi-agent interactions.
\newblock In \emph{2020 IEEE/RSJ International Conference on Intelligent Robots and Systems (IROS)}, pages 6252--6259, 2020.
\newblock \doi{10.1109/IROS45743.2020.9341499}.

\bibitem[Zhou et~al.(2020)Zhou, Gao, Pan, and Shen]{9196996}
Boyu Zhou, Fei Gao, Jie Pan, and Shaojie Shen.
\newblock Robust real-time uav replanning using guided gradient-based optimization and topological paths.
\newblock In \emph{2020 IEEE International Conference on Robotics and Automation (ICRA)}, pages 1208--1214, 2020.
\newblock \doi{10.1109/ICRA40945.2020.9196996}.

\end{thebibliography}

\end{document}